\documentclass{article}
\usepackage{float}

\usepackage[nonatbib,dblblindworkshop, final]{neurips_2026}

\workshoptitle{2nd Workshop on Advances in Representation Learning for Earth Observation}

\usepackage[utf8]{inputenc}      %
\usepackage[T1]{fontenc}         %
\usepackage{hyperref}            %

\newcommand{\appref}[1]{App.~\ref{#1}}
\usepackage{url}                 %
\usepackage{booktabs}            %
\usepackage{amsfonts}            %
\usepackage{nicefrac}            %
\usepackage{microtype}           %
\usepackage{xcolor}              %
\usepackage{graphicx}            %
\usepackage{float}               %
\usepackage[percent]{overpic}    %
\usepackage{siunitx}             %
\usepackage{paralist}
\usepackage[font=small]{caption} %

\usepackage[
  backend=biber,
  style=numeric,
  sorting=none,
  giveninits=true,
  maxnames=15,
  minnames=15,
  natbib=true,
  url=true,
  doi=true,
  isbn=false
]{biblatex}
\DeclareNameAlias{author}{family-given}

\AtEveryBibitem{%
  \clearlist{language}%
  \clearfield{pages}%
  \clearfield{volume}%
  \clearfield{number}%
  \clearfield{issue}%
  \clearfield{month}%
  \clearfield{publisher}%
  \clearfield{location}%
  \clearfield{edition}%
  \clearfield{series}%
  \clearfield{note}%
  \clearfield{urlyear}%
  \clearfield{urlmonth}%
  \clearfield{urlday}%
}

\DeclareBibliographyDriver{article}{%
  \printnames{author}.\space
  \printfield{title}.\space
  \printfield{journaltitle}.\space
  \printdate.\space
  \printfield{doi}.%
  \finentry
}
\DeclareBibliographyDriver{inproceedings}{%
  \printnames{author}.\space
  \printfield{title}.\space
  \printfield{booktitle}.\space
  \printdate.\space
  \printfield{doi}.%
  \finentry
}
\DeclareBibliographyDriver{misc}{%
  \printnames{author}.\space
  \printfield{title}.\space
  \printdate.\space
  \printfield{doi}.\space
  \printfield{url}.%
  \finentry
}
\DeclareBibliographyDriver{online}{%
  \printfield{title}.\space
  \printfield{url}.%
  \finentry
}

\DeclareFieldFormat{doi}{\mkbibacro{DOI}\addcolon\space\href{https://doi.org/#1}{#1}}
\DeclareFieldFormat{url}{\url{#1}}

\hypersetup{pdfauthor={}, pdftitle={}, pdfsubject={}, pdfkeywords={}}

\title{GNN-VIEW: Visual Inspection and Explanation of\\ Graph-based Earth-System Models}

\workshoptitle{2nd Workshop on Advances in Representation Learning for Earth Observation}

\author{\textbf{Alessandro Lucatelli}$^*$ \quad \textbf{Ferdinand Unterhuber} \quad \textbf{Katharina Krenzer} \quad \\ 
\textbf{Mennatallah El-Assady} \quad \textbf{Benedikt Soja} \quad \textbf{Christina Humer}$^*$ \quad \textbf{Joel Oskarsson}$^*$ \quad \\
ETH Zürich, Switzerland \\ 
$^*$\texttt{\{alucatelli|humerc|ojoel\}@ethz.ch}
}

\begin{document}

\maketitle

\begin{abstract}
  Machine learning Earth-system models are becoming increasingly important, and so is their development. Graph-based models, one of the most prominent approaches, allow developers to constrain the learning process before training via graph connectivity. Yet, developing these models remains challenging: the graph is built algorithmically with little intuition about the final result beyond static plotting, which requires extra work and is often not detailed enough. Even after successful training, how the model arrives at forecasts remains obscure. We present GNN-VIEW, an interactive browser-based tool that supports the development of such systems. GNN-VIEW exposes the graph's information pathways, allowing model developers to debug the graph structure before training. After training, GNN-VIEW uses attribution-based explainability to reveal which input features are used by the trained model for a given forecast.
  This helps developers understand \textit{why} a model fails. We showcase the usefulness of GNN-VIEW in the construction stage of a graph for an ocean forecasting model, and in explaining poorly predicted precipitation of a trained weather forecasting model during Hurricane Laura. 
\end{abstract}

\section{Introduction}
\label{sec:intro}

Machine learning (ML) has opened up new possibilities for making good use of Earth Observation (EO) data in Earth-system modeling.
ML models can learn an internal representation of the Earth system either from reanalysis datasets \citep{bodnar_foundation_2025,holmberg_njord_2026,lam_learning_2023} derived from EO data, or by directly ingesting observations to produce forecasts \citep{alexe_graphdop_2024}.
A particularly effective and popular modeling paradigm has been graph-based Earth-system models using Graph Neural Networks (GNNs). Their graph structure encodes the spatial connectivity of the Earth system and creates a spatial inductive bias that constrains the representational capability of the model before training. This is essential for model generalization, and improper graph initialization can degrade forecast capabilities \citep{lam_learning_2023,holmberg_njord_2026}. These models follow an \textit{encoder-processor-decoder} structure: physical states live on an input grid, while the computation happens on a coarser mesh graph. Three sets of edges tie these together: grid-to-mesh edges (G2M) encode inputs into the mesh, mesh-to-mesh edges (M2M) propagate information through message passing, and mesh-to-grid edges (M2G) decode the state back into the grid points (Fig.~\ref{fig:workflow}b).
GNN-based weather models have matched and often surpassed the forecast ability of operational numerical prediction at a fraction of the inference cost \citep{lam_learning_2023, rasp_weatherbench_2024}, and are increasingly adopted operationally \citep{lang_aifs_2024}. The same design and development challenges extend beyond atmospheric modeling to domains such as ocean \citep{holmberg_njord_2026, holmberg_accurate_2025}, wildfire \citep{michail_firecastnet_2026}, hydrology \citep{bentivoglio_multi-scale_2025} and ice sheet \citep{koo_graph_2025} modeling.

Despite its key role in these models, validation of graph construction remains challenging. Developers algorithmically construct large-scale graphs in which structural anomalies hide in raw tensor files and can silently degrade performance. 
To assess graph construction, developers currently rely on custom static plotting of specific graph regions \citep{anemoi_graph_creator,mllam_graphs,holmberg_njord_2026}, which is not scalable and domain-dependent. 
After successful graph initialization, the challenges continue in post-training. Assessing why a model produced a specific forecast is hard: aggregate metrics such as RMSE over a held-out period report \textit{how much} and \textit{where} a model fails, but not \textit{why} the model made the inaccurate forecast. Understanding this is crucial for debugging the model and uncovering potential shortcut learning \citep{geirhos_shortcut_2020}, ultimately enabling the development of more robust and trustworthy models.

In both development challenges, developers need the same: to see what aggregate metrics and raw tensors hide. Visual analytics is an established paradigm for gaining a better understanding of deep learning \citep{hohman_summit_2020, kahng_activis_2018, spinner_innspector_2024}, and GNNs specifically \citep{jin_gnnlens_2022}. 
These systems are model-agnostic and could in theory be applied to Earth-system models. Yet, they do not incorporate the geographic semantics needed here. The two closest works differ significantly: GNNLens~\citep{jin_gnnlens_2022} is for node classification tasks only, and CloudNine~\citep{jeon_observation_2024} addresses the data assimilation stage, rather than the forecast.
Within the Earth-system GNN community, libraries such as Neural-LAM \citep{oskarsson_graph-based_2023} and Anemoi \citep{ecmwf_anemoi-graphs_2024} provide basic plotting utilities, yet they are insufficient to effectively inspect dense, multi-layer large-scale graphs. 

Our contribution addresses this gap. We present GNN-VIEW\footnote{Try GNN-VIEW live demo: \href{https://gnn-view.ivia.ch}{gnn-view.ivia.ch}. Code available: \href{https://github.com/AlucateL/GNN-VIEW}{github.com/AlucateL/GNN-VIEW}
}: an interactive visual tool that supports the full development pipeline of Earth-system GNNs, from graph inspection to forecast explanation, across domains. We demonstrate its usefulness in two scenarios: identifying construction flaws in an ocean forecasting graph and explaining a mispredicted field during Hurricane Laura.

\section{GNN-VIEW}
\label{sec:tool}

\begin{figure}[t]
  \centering
  \includegraphics[width=1\linewidth]{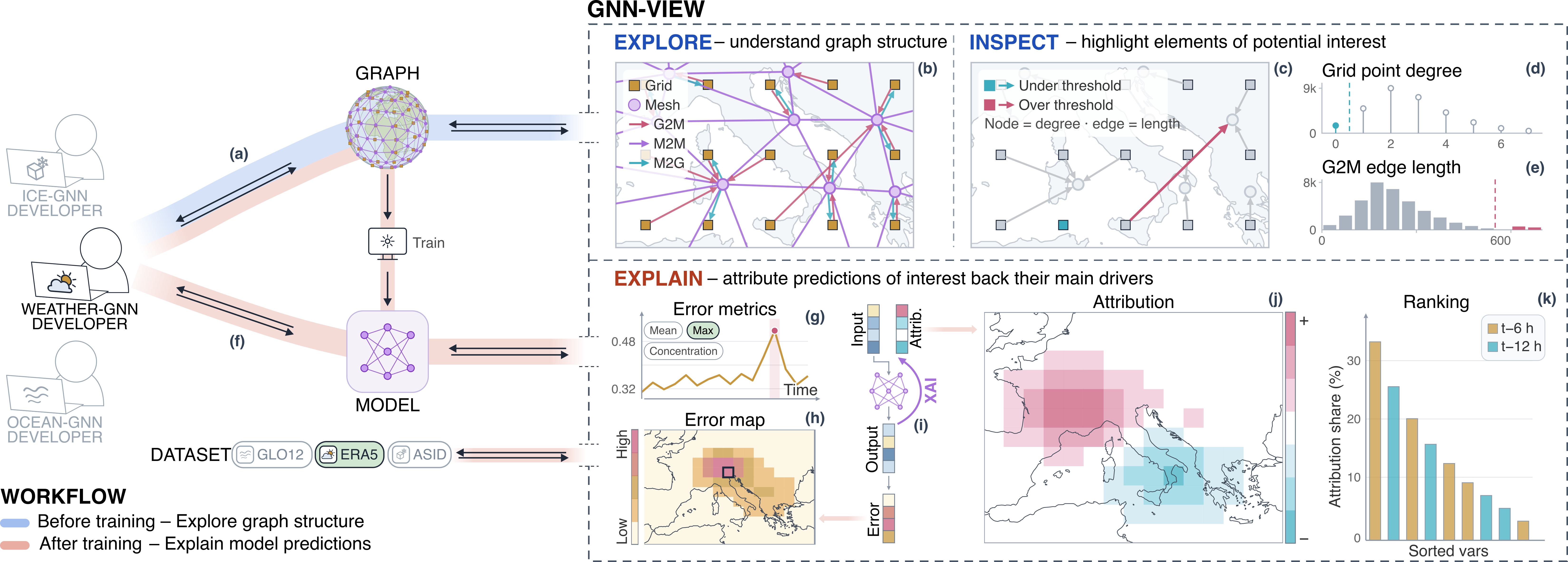}
    \caption{GNN-VIEW follows a developer's workflow. (\textbf{a}) Before training, it allows (\textbf{b}) exploration of the graph structure and its components (grid points, mesh nodes, edges), followed by (\textbf{c}) detailed inspection of the graph structure via (\textbf{d-e}) manual thresholding of graph statistics. (\textbf{f}) After model training, developers can get (\textbf{i-k}) attribution-based explanations for (\textbf{g,h}) predictions of interest to detect what lies behind model failure.}
  \label{fig:workflow}
\end{figure}

GNN-VIEW is organized around three modes that follow a developer's workflow (Fig.~\ref{fig:workflow}) and analyzes model representational capabilities from different angles: \textit{Explore} and \textit{Inspect} describe the information pathways available in the graph; \textit{Explain} measures how the trained model uses them. 

\vspace{-0.8em}
\paragraph{Graph structure exploration} GNN-VIEW allows exploration of the complete encoder-processor-decoder graph over a projected global map via two coordinated modes: \emph{Explore} and \emph{Inspect} (Fig.~\ref{fig:workflow}a-e).
The \textit{Explore} mode shows the full graph (Fig.~\ref{fig:workflow}b), letting the developer build an initial understanding of the graph structure the model operates on. 
The \textit{Inspect} mode enables precise navigation to find mis-specified graph elements. 
We characterize the local structure of a selected subgraph with three quantities: node in- and out-degree, and edge length (Fig.~\ref{fig:workflow}d-e). Developers can specify upper and lower thresholds for these statistics to highlight elements of potential interest in the map (Fig.~\ref{fig:workflow}c). Coupling statistical selection with geographic patterns enables efficient graph diagnostics.

\vspace{-0.8em}
\paragraph{Explaining model predictions} GNN-VIEW supports post-training explainability through its \textit{Explain} mode (Fig.~\ref{fig:workflow}f-k), with a two-stage workflow: the user first identifies a target prediction (time, variable, grid-point) of interest, then the tool attributes it back to its inputs (timesteps, variables, grid-points).
Complementary error metrics resolved over time, space, and variables (Fig.~\ref{fig:workflow}g,h) guide developers toward periods, regions, and variables of poor performance to ultimately help with model debugging.
When a target is selected, the user can explore which inputs contribute to the prediction with attribution-based XAI \citep{wang_gradient_2024} (Fig.~\ref{fig:workflow}i).
To trade off latency against exactness, two attribution methods are available: Input~$\times$~Gradient (IxG) \citep{shrikumar_learning_2017} for a fast qualitative overview and Integrated Gradients (IG) \citep{sundararajan_axiomatic_2017} for slower and more accurate attributions.
Both methods produce signed scores per input feature: positive values indicate features that increase the model's prediction above a baseline value, while negative values indicate those that decrease it. For a more detailed description of IG completeness and sanity checks~\citep{adebayo_sanity_2018}, see \appref{app:ig} and \appref{app:sanity}, respectively.
Summing absolute attribution scores over space and normalizing by the total gives the \emph{attribution share} of each input field: how much each input (timestep, variable) contributes to the total attribution.

Resulting attributions are extremely high-dimensional --- one tensor (timesteps, variables, grid-points) for each target --- and presenting them to the user in a meaningful way is challenging. We tackle this as follows: Attribution scores are shown on a heatmap (Fig.~\ref{fig:workflow}j) to highlight their spatial distribution. We show two of these side by side to compare spatial structures across input timesteps or variables. Additionally, a ranked list of input variables, ordered by attribution share (Fig.~\ref{fig:workflow}k), averages the spatial dimension out and informs about high-level model behavior over time and variables.

\vspace{-0.8em}
\paragraph{Technical setup} GNN-VIEW operates on three independent entities, each loaded separately: \textit{Explore} and \textit{Inspect} only need the graph topology (Fig.~\ref{fig:workflow}a). \textit{Explain} additionally needs the trained model and dataset (Fig.~\ref{fig:workflow}f). The graph topology can be uploaded in either Neural-LAM \citep{oskarsson_graph-based_2023} or Anemoi \citep{lang_aifs_2024, ecmwf_anemoi-graphs_2024} format, for both flat \citep{lam_learning_2023,lang_aifs_2024} and hierarchical graphs \citep{oskarsson_probabilistic_2024, holmberg_njord_2026}. No assumptions about mesh generation are made: any graph can be visualized regardless of how it was built. Tool latency scales linearly with the number of graph elements (\appref{app:latency}). The model is uploaded as a TorchScript .pt file.
The tool ships with two default datasets that users can select: ERA5 \citep{hersbach_era5_2020} at 1.5\textdegree~and 6-hourly resolution \citep{rasp_weatherbench_2024} for weather modeling, and an adaptation of GLORYS12 \citep{el_aouni_oceanbench_2026}, re-gridded to 0.25\textdegree~and with sea-ice added, for oceanography modeling. Local deployment is also possible (\appref{app:deployment}).

\section{Use cases: GNN-VIEW in action}
\label{sec:usecase}

\paragraph{Spotting flaws in a graph}
\label{sec:usecase_1}

\begin{figure}[t]
  \centering
  \includegraphics[width=\linewidth]{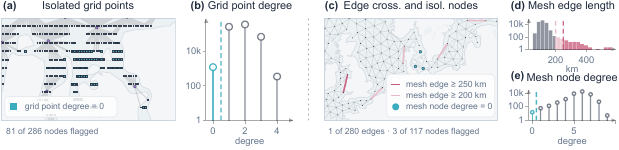}
    \caption{GNN-VIEW Inspect mode, on the Njord ocean graph. (\textbf{a-b}) Grid points thresholded on out-degree in the encoder subgraph (G2M). (\textbf{c-e}) Mesh edges thresholded on length and mesh nodes thresholded on degree. All distributions span the full graph, not just the viewport in the figure.}

  \label{fig:inspect}
\end{figure}

We showcase the use of GNN-VIEW for validating graph structures on the graph used for the Njord ocean forecasting model \citep{holmberg_njord_2026}. Its mesh is generated by spatially clustering sea surface grid points, so the number of clusters (i.e., mesh nodes) is a free parameter. 
Ocean graphs are challenging to build because of the irregular domain they operate on. This implies several, often competing, requirements: edges should not cross major land masses, yet still cover all water basins, and every sea grid point should be connected to the mesh \citep{holmberg_njord_2026,cuervo-londono_voronoi-induced_2025}. To this end, Njord's developers cap edge lengths above a free threshold parameter and reconnect isolated elements to the mesh via a repair pass \citep{holmberg_njord_2026}. 
Setting the free parameters is iterative, time-consuming, and error-prone. 

Fig.~\ref{fig:inspect} illustrates how \textit{Inspect} mode can help inform this process. 
By setting the threshold for out-degree to 0 in the G2M subgraph (Fig.~\ref{fig:inspect}b), developers can surface isolated points whose states are not encoded in the model (Fig.~\ref{fig:inspect}a), here undermining the correct representation of the Kara Sea. This leads to a more informed decision on whether to run a disconnection repair algorithm or increase the number of mesh nodes. 
In addition, by thresholding the mesh edge length and mesh node degree, two other failure modes can be seen: a mesh edge crosses over Wales, connecting water masses separated by land, and mesh nodes on the western shores of Denmark are isolated, leaving the Baltic Sea disconnected from the rest of the ocean (Fig.~\ref{fig:inspect}c-e). This informs developers about the edge length cap to use and the need to reconnect the nodes to the mesh. Fig.~\ref{fig:inspect}c contrasts two cutoffs: 250 km isolates the land-crossing edge, while 200 km also captures legitimate long-range links. GNN-VIEW makes this sensitivity visible to calibrate the threshold effectively.

\vspace{-0.8em}
\paragraph{Explaining a hurricane mis-prediction}
\label{sec:usecase_2}

We illustrate the \textit{Explain} mode using a GraphCast-style GNN \citep{lam_learning_2023,oskarsson_probabilistic_2024} trained on 1.5\textdegree~ERA5 data, focusing on forecasts of Hurricane Laura in 2020. The target time is 2020-08-27 12:00~UTC, i.e. six hours after Hurricane Laura made landfall at Category~4 intensity. 
The model takes as input two past atmospheric states (00:00 and 06:00 UTC), for which the storm is at peak intensity. Hence it is asked to predict its collapse as it traverses from ocean to land.

\begin{figure}[t]
  \centering
  \includegraphics[width=\linewidth]{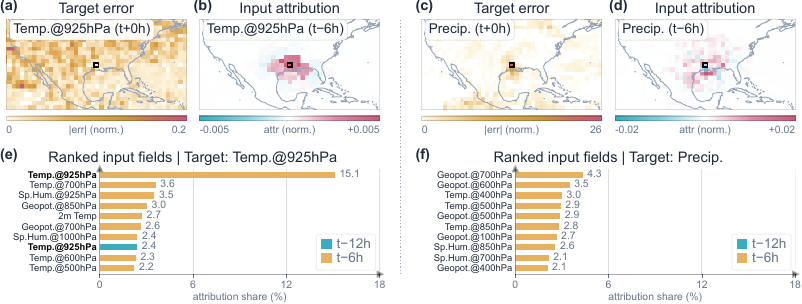}
    \caption{GNN-VIEW Explain mode on a GraphCast-style GNN.
    (\emph{Black square}) Target node 31.5\textdegree~N 93.0\textdegree~W. 
    XAI target field is (\emph{Left}) Temperature at 925 hPa and (\emph{Right}) Total 6-hourly Precipitation. 
    (\textbf{a,c}) Target error: absolute standardized error heatmap of the target field.
    (\textbf{b,d}) Input attribution: signed IG attribution of the target node's prediction w.r.t. that field at t$-$6~h, scaled to each panel's own largest absolute attribution excluding the target node. (\textbf{e,f}) Ranked input fields: top-10 input features by IG attribution share.
    }
  \label{fig:explain}
\end{figure}

In Fig.~\ref{fig:explain} we compare the Explain mode attribution patterns of the accurately predicted temperature at 925 hPa (Fig.~\ref{fig:explain}a) with the poorly predicted 6-hour precipitation (Fig.~\ref{fig:explain}c).
For temperature, the attribution ranking (Fig.~\ref{fig:explain}e) is dominated by a single input: the same field at t$-$6~h; and the attribution is compact and centered around the target point (Fig.~\ref{fig:explain}b). The model uses the recent local state for the prediction.
For precipitation, the pattern is qualitatively different. The attribution ranking is relatively flat (Fig.~\ref{fig:explain}f), led by geopotential and temperature at higher altitudes, which characterize the surroundings rather than the storm itself. Known physical drivers of precipitation, such as moisture and winds, enter the ranking marginally or not at all. Precipitation itself at t$-$6~h does not enter the top ten, and its attribution map is weak and mixed in sign (Fig.~\ref{fig:explain}d).
One hypothesis consistent with
this pattern is that the model at this coarse resolution falls back to correlating precipitation with larger-scale circulation patterns rather than drawing on relevant physical processes.
This is exactly the kind of testable hypothesis GNN-VIEW is built to generate.

\section{Summary and future work}
\label{sec:conclusion}

Developing accurate graph-based Earth-system models with good representational capabilities requires the validation of the graph structure \textit{before training}, and debugging failure modes of model predictions \textit{after training}. GNN-VIEW is an interactive specialized visualization tool that brings these into developers' view.
Through two scenarios, a poorly connected ocean graph and a mispredicted precipitation field, we have showcased how this tool can give Earth-model developers actionable insights to debug and improve their models. In both use cases GNN-VIEW returns interactive evidence, not verdicts, enabling
an interactive loop with the tool (Fig.~\ref{fig:workflow}a,f) that ultimately allows for a better and more transparent development.
Some design choices behind GNN-VIEW come with limitations that can be explored in future work.
Structure-aware explainers \citep{luo_parameterized_2020} could connect model behavior closer to the graph structure. Explainability of forecast rollouts \citep{essafouri_explanation_2026} and probabilistic model ensembles would also broaden the tool's impact in the Earth-system modeling community. Further evaluation of the tool's impact on developers' workflow through user studies would also be valuable.

\begin{ack}
This research was supported by the ETH AI Center through ETH AI Center postdoctoral fellowships to Christina Humer and Joel Oskarsson and by the International Geneva AI Innovation Programme (IGAIP) with funding from the Swiss Foreign Ministry.
\end{ack}

\printbibliography

@article{jeon_observation_2024,
	title = {Observation impact explanation in atmospheric state estimation using hierarchical message-passing graph neural networks},
	volume = {5},
	issn = {2632-2153},
	url = {https://doi.org/10.1088/2632-2153/ad8981},
	doi = {10.1088/2632-2153/ad8981},
	language = {en},
	number = {4},
	urldate = {2026-08-04},
	journal = {Mach. Learn.: Sci. Technol.},
	publisher = {IOP Publishing},
	author = {Jeon, Hyeon-Ju and Kang, Jeon-ho and Kwon, In-Hyuk and Lee, O-Joun},
	month = nov,
	year = {2024},
	pages = {045036}
}

@misc{alexe_graphdop_2024,
	title = {{GraphDOP}: {Towards} skilful data-driven medium-range weather forecasts learnt and initialised directly from observations},
	shorttitle = {{GraphDOP}},
	url = {http://arxiv.org/abs/2412.15687},
	doi = {10.48550/arXiv.2412.15687},
	publisher = {arXiv},
	author = {Alexe, Mihai and Boucher, Eulalie and Lean, Peter and Pinnington, Ewan and Laloyaux, Patrick and McNally, Anthony and Lang, Simon and Chantry, Matthew and Burrows, Chris and Chrust, Marcin and Pinault, Florian and Villeneuve, Ethel and Bormann, Niels and Healy, Sean},
	month = dec,
	year = {2024},
	note = {arXiv:2412.15687 [physics.ao-ph]}
}

@article{bentivoglio_multi-scale_2025,
	title = {Multi-scale hydraulic graph neural networks for flood modelling},
	volume = {25},
	issn = {1561-8633},
	url = {https://nhess.copernicus.org/articles/25/335/2025/},
	doi = {10.5194/nhess-25-335-2025},
	number = {1},
	journal = {Natural Hazards and Earth System Sciences},
	publisher = {Copernicus GmbH},
	author = {Bentivoglio, Roberto and Isufi, Elvin and Jonkman, Sebastiaan Nicolas and Taormina, Riccardo},
	month = jan,
	year = {2025},
	pages = {335--351}
}

@article{bodnar_foundation_2025,
	title = {A foundation model for the {Earth} system},
	volume = {641},
	issn = {1476-4687},
	url = {https://www.nature.com/articles/s41586-025-09005-y},
	doi = {10.1038/s41586-025-09005-y},
	number = {8065},
	journal = {Nature},
	publisher = {Nature Publishing Group},
	author = {Bodnar, Cristian and Bruinsma, Wessel P. and Lucic, Ana and Stanley, Megan and Allen, Anna and Brandstetter, Johannes and Garvan, Patrick and Riechert, Maik and Weyn, Jonathan A. and Dong, Haiyu and Gupta, Jayesh K. and Thambiratnam, Kit and Archibald, Alexander T. and Wu, Chun-Chieh and Heider, Elizabeth and Welling, Max and Turner, Richard E. and Perdikaris, Paris},
	month = may,
	year = {2025},
	pages = {1180--1187}
}

@article{cuervo-londono_voronoi-induced_2025,
	title = {Voronoi-{Induced} {Artifacts} from {Grid}-to-{Mesh} {Coupling} and {Bathymetry}-{Aware} {Meshes} in {Graph} {Neural} {Networks} for {Sea} {Surface} {Temperature} {Forecasting}},
	volume = {14},
	issn = {2079-9292},
	url = {https://www.mdpi.com/2079-9292/14/24/4841},
	doi = {10.3390/electronics14244841},
	number = {24},
	journal = {Electronics},
	publisher = {Multidisciplinary Digital Publishing Institute},
	author = {Cuervo-Londoño, Giovanny A. and Reyes, José G. and Rodríguez-Santana, Ángel and Sánchez, Javier},
	month = jan,
	year = {2025},
	pages = {4841}
}

@misc{ecmwf_anemoi-graphs_2024,
	title = {anemoi-graphs: {Graph} construction for data-driven weather models},
	url = {https://github.com/ecmwf/anemoi-core},
	author = {{ECMWF}},
	year = {2024}
}

@article{el_aouni_oceanbench_2026,
	title = {{OceanBench}: {A} {Benchmark} for {Data}-{Driven} {Global} {Ocean} {Forecasting} systems},
	volume = {38},
	url = {https://proceedings.neurips.cc/paper_files/paper/2025/hash/0d1635dfa43efe7477f9b6dab3168fd8-Abstract-Datasets_and_Benchmarks_Track.html},
	doi = {10.52202/085713-0303},
	journal = {Advances in Neural Information Processing Systems},
	author = {El Aouni, Anass and Gaudel, Quentin and Johnson, J. Emmanuel and Charly, Regnier and Le Sommer, Julien and van Gennip, Simon and Fablet, Ronan and Drevillon, Marie and Drillet, Yann and Le Traon, Pierre},
	month = apr,
	year = {2026}
}

@misc{essafouri_explanation_2026,
	title = {Explanation of {Dynamic} {Physical} {Field} {Predictions} using {WassersteinGrad}: {Application} to {Autoregressive} {Weather} {Forecasting}},
	url = {http://arxiv.org/abs/2604.22580},
	doi = {10.48550/arXiv.2604.22580},
	publisher = {arXiv},
	author = {Essafouri, Younes and Raynaud, Laure and Drozda, Luciano and Risser, Laurent},
	month = apr,
	year = {2026},
	note = {arXiv:2604.22580 [stat.ML]}
}

@article{geirhos_shortcut_2020,
	title = {Shortcut learning in deep neural networks},
	volume = {2},
	issn = {2522-5839},
	url = {https://www.nature.com/articles/s42256-020-00257-z},
	doi = {10.1038/s42256-020-00257-z},
	number = {11},
	journal = {Nature Machine Intelligence},
	publisher = {Nature Publishing Group},
	author = {Geirhos, Robert and Jacobsen, Jörn-Henrik and Michaelis, Claudio and Zemel, Richard and Brendel, Wieland and Bethge, Matthias and Wichmann, Felix A.},
	month = nov,
	year = {2020},
	pages = {665--673}
}

@article{hersbach_era5_2020,
	title = {The {ERA5} global reanalysis},
	volume = {146},
	issn = {1477-870X},
	url = {https://onlinelibrary.wiley.com/doi/abs/10.1002/qj.3803},
	doi = {10.1002/qj.3803},
	number = {730},
	journal = {Quarterly Journal of the Royal Meteorological Society},
	author = {Hersbach, Hans and Bell, Bill and Berrisford, Paul and Hirahara, Shoji and Horányi, András and Muñoz-Sabater, Joaquín and Nicolas, Julien and Peubey, Carole and Radu, Raluca and Schepers, Dinand and Simmons, Adrian and Soci, Cornel and Abdalla, Saleh and Abellan, Xavier and Balsamo, Gianpaolo and Bechtold, Peter and Biavati, Gionata and Bidlot, Jean and Bonavita, Massimo and De Chiara, Giovanna and Dahlgren, Per and Dee, Dick and Diamantakis, Michail and Dragani, Rossana and Flemming, Johannes and Forbes, Richard and Fuentes, Manuel and Geer, Alan and Haimberger, Leo and Healy, Sean and Hogan, Robin J. and Hólm, Elías and Janisková, Marta and Keeley, Sarah and Laloyaux, Patrick and Lopez, Philippe and Lupu, Cristina and Radnoti, Gabor and de Rosnay, Patricia and Rozum, Iryna and Vamborg, Freja and Villaume, Sebastien and Thépaut, Jean-Noël},
	year = {2020},
	pages = {1999--2049}
}

@article{hohman_summit_2020,
	title = {Summit: {Scaling} {Deep} {Learning} {Interpretability} by {Visualizing} {Activation} and {Attribution} {Summarizations}},
	volume = {26},
	doi = {10.1109/TVCG.2019.2934659},
	number = {1},
	journal = {IEEE Transactions on Visualization and Computer Graphics},
	author = {Hohman, Fred and Park, Haekyu and Robinson, Caleb and Chau, Duen Horng},
	year = {2020},
	pages = {1096--1106}
}

@article{holmberg_accurate_2025,
	title = {Accurate {Mediterranean} {Sea} forecasting via graph-based deep learning},
	volume = {15},
	issn = {2045-2322},
	url = {https://www.nature.com/articles/s41598-025-31177-w},
	doi = {10.1038/s41598-025-31177-w},
	number = {1},
	journal = {Scientific Reports},
	publisher = {Nature Publishing Group},
	author = {Holmberg, Daniel and Clementi, Emanuela and Epicoco, Italo and Roos, Teemu},
	month = dec,
	year = {2025},
	pages = {45051}
}

@misc{holmberg_njord_2026,
	title = {Njord: {A} {Probabilistic} {Graph} {Neural} {Network} for {Ensemble} {Ocean} {Forecasting}},
	shorttitle = {Njord},
	url = {http://arxiv.org/abs/2605.15470},
	doi = {10.48550/arXiv.2605.15470},
	publisher = {arXiv},
	author = {Holmberg, Daniel and Oskarsson, Joel and Wikingsson, Erik and Lindsten, Fredrik and Roos, Teemu},
	month = may,
	year = {2026},
	note = {arXiv:2605.15470 [cs.LG]}
}

@article{jin_gnnlens_2022,
	title = {{GNNLens}: {A} {Visual} {Analytics} {Approach} for {Prediction} {Error} {Diagnosis} of {Graph} {Neural} {Networks}},
	shorttitle = {{GNNLens}},
	volume = {29},
	issn = {1941-0506},
	url = {https://ieeexplore.ieee.org/document/9705076},
	doi = {10.1109/TVCG.2022.3148107},
	number = {6},
	journal = {IEEE Transactions on Visualization and Computer Graphics},
	author = {Jin, Zhihua and Wang, Yong and Wang, Qianwen and Ming, Yao and Ma, Tengfei and Qu, Huamin},
	month = jun,
	year = {2023},
	pages = {3024--3038}
}

@article{kahng_activis_2018,
	title = {{ActiVis}: {Visual} {Exploration} of {Industry}-{Scale} {Deep} {Neural} {Network} {Models}},
	volume = {24},
	doi = {10.1109/TVCG.2017.2744718},
	number = {1},
	journal = {IEEE Transactions on Visualization and Computer Graphics},
	author = {Kahng, Minsuk and Andrews, Pierre Y. and Kalro, Aditya and Chau, Duen Horng},
	year = {2018},
	pages = {88--97}
}

@article{koo_graph_2025,
	title = {Graph convolutional network as a fast statistical emulator for numerical ice sheet modeling},
	volume = {71},
	issn = {0022-1430, 1727-5652},
	url = {https://www.cambridge.org/core/journals/journal-of-glaciology/article/graph-convolutional-network-as-a-fast-statistical-emulator-for-numerical-ice-sheet-modeling/3BF933067DFFB16949CDADBFE95086AB},
	doi = {10.1017/jog.2024.93},
	journal = {Journal of Glaciology},
	author = {Koo, Younghyun and Rahnemoonfar, Maryam},
	month = jan,
	year = {2025},
	pages = {e15}
}

@article{lam_learning_2023,
	title = {Learning skillful medium-range global weather forecasting},
	volume = {382},
	url = {https://www.science.org/doi/10.1126/science.adi2336},
	doi = {10.1126/science.adi2336},
	number = {6677},
	journal = {Science},
	publisher = {American Association for the Advancement of Science},
	author = {Lam, Remi and Sanchez-Gonzalez, Alvaro and Willson, Matthew and Wirnsberger, Peter and Fortunato, Meire and Alet, Ferran and Ravuri, Suman and Ewalds, Timo and Eaton-Rosen, Zach and Hu, Weihua and Merose, Alexander and Hoyer, Stephan and Holland, George and Vinyals, Oriol and Stott, Jacklynn and Pritzel, Alexander and Mohamed, Shakir and Battaglia, Peter},
	month = dec,
	year = {2023},
	pages = {1416--1421}
}

@misc{lang_aifs_2024,
	title = {{AIFS} -- {ECMWF}'s data-driven forecasting system},
	url = {http://arxiv.org/abs/2406.01465},
	doi = {10.48550/arXiv.2406.01465},
	publisher = {arXiv},
	author = {Lang, Simon and Alexe, Mihai and Chantry, Matthew and Dramsch, Jesper and Pinault, Florian and Raoult, Baudouin and Clare, Mariana C. A. and Lessig, Christian and Maier-Gerber, Michael and Magnusson, Linus and Bouallègue, Zied Ben and Nemesio, Ana Prieto and Dueben, Peter D. and Brown, Andrew and Pappenberger, Florian and Rabier, Florence},
	month = aug,
	year = {2024},
	note = {arXiv:2406.01465 [physics.ao-ph]}
}

@inproceedings{luo_parameterized_2020,
	address = {Red Hook, NY, USA},
	series = {{NIPS} '20},
	title = {Parameterized explainer for graph neural network},
	isbn = {978-1-7138-2954-6},
	url = {https://dl.acm.org/doi/10.5555/3495724.3497370},
	booktitle = {Proceedings of the 34th {International} {Conference} on {Neural} {Information} {Processing} {Systems}},
	publisher = {Curran Associates Inc.},
	author = {Luo, Dongsheng and Cheng, Wei and Xu, Dongkuan and Yu, Wenchao and Zong, Bo and Chen, Haifeng and Zhang, Xiang},
	month = dec,
	year = {2020},
	pages = {19620--19631}
}

@inproceedings{oskarsson_graph-based_2023,
  title = {Graph-based Neural Weather Prediction for Limited Area Modeling},
  author = {Oskarsson, Joel and Landelius, Tomas and Lindsten, Fredrik},
  booktitle = {NeurIPS 2023 Workshop on Tackling Climate Change with Machine Learning},
  year = {2023},
}

@inproceedings{oskarsson_probabilistic_2024,
  title = {Probabilistic Weather Forecasting with Hierarchical Graph Neural Networks},
  author = {Oskarsson, Joel and Landelius, Tomas and Deisenroth, Marc Peter and Lindsten, Fredrik},
  year = {2024},
  booktitle = {Advances in Neural Information Processing Systems},
  volume = {37},
}

@article{rasp_weatherbench_2024,
	title = {{WeatherBench} 2: {A} {Benchmark} for the {Next} {Generation} of {Data}-{Driven} {Global} {Weather} {Models}},
	shorttitle = {{WeatherBench} 2},
	volume = {16},
	issn = {1942-2466},
	url = {https://agupubs.onlinelibrary.wiley.com/doi/abs/10.1029/2023MS004019},
	doi = {10.1029/2023MS004019},
	number = {6},
	journal = {Journal of Advances in Modeling Earth Systems},
	author = {Rasp, Stephan and Hoyer, Stephan and Merose, Alexander and Langmore, Ian and Battaglia, Peter and Russell, Tyler and Sanchez-Gonzalez, Alvaro and Yang, Vivian and Carver, Rob and Agrawal, Shreya and Chantry, Matthew and Ben Bouallegue, Zied and Dueben, Peter and Bromberg, Carla and Sisk, Jared and Barrington, Luke and Bell, Aaron and Sha, Fei},
	month = jun,
	year = {2024},
	pages = {e2023MS004019}
}

@inproceedings{shrikumar_learning_2017,
	address = {Sydney, NSW, Australia},
	series = {{ICML}'17},
	title = {Learning important features through propagating activation differences},
	url = {https://dl.acm.org/doi/10.5555/3305890.3306006},
	booktitle = {Proceedings of the 34th {International} {Conference} on {Machine} {Learning} - {Volume} 70},
	publisher = {JMLR.org},
	author = {Shrikumar, Avanti and Greenside, Peyton and Kundaje, Anshul},
	month = aug,
	year = {2017},
	pages = {3145--3153}
}

@misc{spinner_innspector_2024,
	title = {{iNNspector}: {Visual}, {Interactive} {Deep} {Model} {Debugging}},
	shorttitle = {{iNNspector}},
	url = {http://arxiv.org/abs/2407.17998},
	doi = {10.48550/arXiv.2407.17998},
	publisher = {arXiv},
	author = {Spinner, Thilo and Fürst, Daniel and El-Assady, Mennatallah},
	month = jul,
	year = {2024},
	note = {arXiv:2407.17998 [cs.HC]}
}

@inproceedings{sundararajan_axiomatic_2017,
	title = {Axiomatic {Attribution} for {Deep} {Networks}},
	issn = {2640-3498},
	url = {https://proceedings.mlr.press/v70/sundararajan17a.html},
	booktitle = {Proceedings of the 34th {International} {Conference} on {Machine} {Learning}},
	publisher = {PMLR},
	author = {Sundararajan, Mukund and Taly, Ankur and Yan, Qiqi},
	month = jul,
	year = {2017},
	pages = {3319--3328}
}

@misc{wang_gradient_2024,
	title = {Gradient based {Feature} {Attribution} in {Explainable} {AI}: {A} {Technical} {Review}},
	shorttitle = {Gradient based {Feature} {Attribution} in {Explainable} {AI}},
	url = {http://arxiv.org/abs/2403.10415},
	doi = {10.48550/arXiv.2403.10415},
	publisher = {arXiv},
	author = {Wang, Yongjie and Zhang, Tong and Guo, Xu and Shen, Zhiqi},
	month = mar,
	year = {2024},
	note = {arXiv:2403.10415 [cs.AI]}
}

@online{mllam_graphs,
	title = {Creating the graph — {Graphs} for data-driven weather models},
	url = {https://mllam.github.io/weather-model-graphs/creating_the_graph.html},
	urldate = {2026-09-01},
    author={{MLLAM} developers}
}

@online{anemoi_graph_creator,
	title = {Graph {Creator} — {Anemoi} {Graphs} 0.1.dev103+gccae142 documentation},
	url = {https://anemoi.readthedocs.io/projects/graphs/en/stable/modules/graph_creator.html},
	urldate = {2026-09-01},
    author={{Anemoi developers}}
}

@article{michail_firecastnet_2026,
	title = {{FireCastNet}: earth-as-a-graph for seasonal fire prediction},
	volume = {16},
	issn = {2045-2322},
	url = {https://doi.org/10.1038/s41598-025-30645-7},
	doi = {10.1038/s41598-025-30645-7},
	journal = {Scientific Reports},
	publisher = {Nature Publishing Group},
	author = {Michail, D. and Davalas, C. and Chafis, K. and others},
	year = {2026},
	pages = {1006}
}

@inproceedings{adebayo_sanity_2018,
	address = {Red Hook, NY, USA},
	series = {{NeurIPS} '18},
	title = {Sanity checks for saliency maps},
	url = {https://dl.acm.org/doi/10.5555/3327546.3327621},
	urldate = {2026-10-10},
	booktitle = {Proceedings of the 32nd {International} {Conference} on {Neural} {Information} {Processing} {Systems}},
	publisher = {Curran Associates Inc.},
	author = {Adebayo, Julius and Gilmer, Justin and Muelly, Michael and Goodfellow, Ian and Hardt, Moritz and Kim, Been},
	month = dec,
	year = {2018},
	pages = {9525--9536}
}

@incollection{hedstrom_fresh_2024,
	title = {A {Fresh} {Look} at {Sanity} {Checks} for {Saliency} {Maps}},
	volume = {2153},
	url = {https://doi.org/10.1007/978-3-031-63787-2_21},
	doi = {10.1007/978-3-031-63787-2_21},
	booktitle = {Explainable Artificial Intelligence},
	series = {Communications in Computer and Information Science},
	publisher = {Springer},
	address = {Cham},
	author = {Hedstr{\"o}m, Anna and Weber, Leander and Lapuschkin, Sebastian and H{\"o}hne, Marina},
	editor = {Longo, Luca and Lapuschkin, Sebastian and Seifert, Christin},
	year = {2024}
}

@article{samek_evaluating_2017,
	title = {Evaluating the {Visualization} of {What} a {Deep} {Neural} {Network} {Has} {Learned}},
	volume = {28},
	url = {https://ieeexplore.ieee.org/document/7552539},
	doi = {10.1109/TNNLS.2016.2599820},
	number = {11},
	journal = {IEEE Transactions on Neural Networks and Learning Systems},
	author = {Samek, Wojciech and Binder, Alexander and Montavon, Gr{\"e}goire and Lapuschkin, Sebastian and M{\"u}ller, Klaus-Robert},
	month = nov,
	year = {2017},
	pages = {2660--2673}
}

@misc{petsiuk_rise_2018,
	title = {{RISE}: {Randomized} {Input} {Sampling} for {Explanation} of {Black}-box {Models}},
	url = {http://arxiv.org/abs/1806.07421},
	doi = {10.48550/arXiv.1806.07421},
	publisher = {arXiv},
	author = {Petsiuk, Vitali and Das, Abir and Saenko, Kate},
	month = jun,
	year = {2018},
	note = {arXiv:1806.07421 [cs.CV]}
}

@article{bommer_finding_2024,
	title = {Finding the {Right} {XAI} {Method}---{A} {Guide} for the {Evaluation} and {Ranking} of {Explainable} {AI} {Methods} in {Climate} {Science}},
	volume = {3},
	url = {https://doi.org/10.1175/AIES-D-23-0074.1},
	doi = {10.1175/AIES-D-23-0074.1},
	number = {3},
	journal = {Artificial Intelligence for the Earth Systems},
	author = {Bommer, Philine L. and Kretschmer, Marlene and Hedstr{\"o}m, Anna and Bareeva, Dilyara and H{\"o}hne, Marina M.-C.},
	month = jun,
	year = {2024},
	pages = {e230074}
}

@article{bach_pixel-wise_2015,
	title = {On {Pixel}-{Wise} {Explanations} for {Non}-{Linear} {Classifier} {Decisions} by {Layer}-{Wise} {Relevance} {Propagation}},
	volume = {10},
	url = {https://doi.org/10.1371/journal.pone.0130140},
	doi = {10.1371/journal.pone.0130140},
	number = {7},
	journal = {PLOS ONE},
	publisher = {Public Library of Science},
	author = {Bach, Sebastian and Binder, Alexander and Montavon, Gr{\"e}goire and Klauschen, Frederick and M{\"u}ller, Klaus-Robert and Samek, Wojciech},
	month = jul,
	year = {2015},
	pages = {e0130140}
}

@inproceedings{bhatt_evaluating_2020,
	address = {Yokohama, Japan},
	title = {Evaluating and aggregating feature-based model explanations},
	url = {https://dl.acm.org/doi/abs/10.5555/3491440.3491857},
	urldate = {2026-10-10},
	doi = {10.24963/ijcai.2020/417},
	booktitle = {Proceedings of the Twenty-Ninth {International} {Joint} {Conference} on {Artificial} {Intelligence}},
	author = {Bhatt, Umang and Weller, Adrian and Moura, Jos{\'e} M. F.},
	year = {2020},
	pages = {3016--3022}
}

@incollection{wickstrom_flexibility_2025,
	title = {From Flexibility to Manipulation: The Slippery Slope of XAI Evaluation},
	url = {https://doi.org/10.1007/978-3-031-92648-8_15},
	urldate = {2026-10-11},
	doi = {10.1007/978-3-031-92648-8_15},
	booktitle = {Computer Vision -- ECCV 2024 Workshops},
	series = {Lecture Notes in Computer Science},
	volume = {15643},
	publisher = {Springer},
	address = {Cham},
	author = {Wickstr{\o}m, K. and H{\"o}hne, M. and Hedstr{\"o}m, A.},
	editor = {Del Bue, A. and Canton, C. and Pont-Tuset, J. and Tommasi, T.},
	year = {2025}
}

\medskip
{
\small
}

\newpage
\appendix

\section{GNN-VIEW details}
\subsection{Single-step inference and evaluation metric} 
\label{app:inference}

Given two consecutive normalized input states (t$-$12~h and t$-$6~h) and a forcing tensor, the model produces the state at $t$. The per-grid-point prediction error is computed on demand for the selected timestep and is rendered as the error heatmap. The single-step timeline summaries (mean and maximum RMSE, plus error concentration) are precomputed over the full test period, so the overview is available without re-running inference.

\subsection{Local deployment}
\label{app:deployment}
The tool can be deployed locally, in which case users can point GNN-VIEW at their own datasets. Inference, error and attribution pipelines stay unchanged, since nothing downstream of loading assumes a particular dataset.

\section{GNN-VIEW latency}
\label{app:latency}

\begin{figure}[tbph]
  \centering
  \includegraphics[width=1\linewidth]{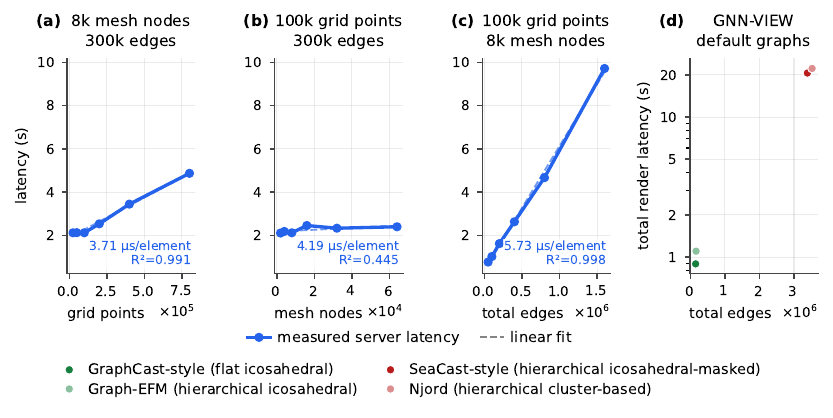}
    \caption{Server latency measured for (\textbf{a-c}) one component at a time, and (\textbf{d}) for real graphs. (\textbf{a}) Latency of rendering an increasing number of grid points; mesh nodes and edges count kept fixed. (\textbf{b}) Latency of rendering an increasing number of mesh nodes; grid-points and edge count kept fixed. (\textbf{c}) Latency of rendering an increasing number of edges; grid-point and mesh-node count kept fixed. (\textbf{d}) Latency of rendering the default graphs available in GNN-VIEW: GraphCast-style and Graph-EFM graphs for weather modeling from \citep{lam_learning_2023,oskarsson_probabilistic_2024} and SeaCast-style and Njord graphs for ocean modeling from \citep{holmberg_accurate_2025,holmberg_njord_2026}.}
  \label{fig:latency}
\end{figure}

We measured server latency on synthetic (Fig.~\ref{fig:latency}a-c), as well as on real graphs (Fig.~\ref{fig:latency}d), on a standard consumer laptop. Synthetic graphs have been generated by plotting an arbitrary number of grid-points, mesh nodes and edges. To measure individual contributions to latency, we increase the number of each one of these elements at a time while keeping the other two fixed (i.e. total edges increase in Fig.~\ref{fig:latency}c, while grid points and mesh nodes are fixed). 

Fig.~\ref{fig:latency}a-c shows that latency scales linearly with the number of components and that it is dominated by edge rendering: 5.73 $\mu$s/element against the 3.71 and 4.19 $\mu$s/element of grid-points and mesh nodes. Edges dominate latency also because they are typically far more numerous than grid-points or mesh nodes in real graphs. For this reason Fig.~\ref{fig:latency}d compares real graphs based on their total number of edges.

Fig.~\ref{fig:latency}d shows the server latency for rendering the graph provided as defaults in GNN-VIEW, taken from recent literature in graph-based weather \citep{oskarsson_probabilistic_2024,lam_learning_2023} and ocean \citep{holmberg_njord_2026,holmberg_accurate_2025} modeling. Latency scales linearly for real graphs too (Fig.~\ref{fig:latency}d): an oceanography graph with approximately $22\times$ more edges requires almost $20 \times$ more time.

Latency measured in Fig.~\ref{fig:latency} is the total latency to render the full graph on screen at once. Typically GNN-VIEW operates on much finer zoom levels where only a portion of interest of the graph is rendered at a time. Given the linear scaling of the latency, this results in a considerably lower latency in common use cases of GNN-VIEW.

\section{Sanity Checks}
\label{app:sanity}

We performed two types of sanity checks to quantify the relationship between explanations and the model: randomization of model parameters \citep{adebayo_sanity_2018} and deletion tests \citep{bach_pixel-wise_2015,samek_evaluating_2017,petsiuk_rise_2018}. This allows us to assess the ``\emph{sensitivity}'' \citep{hedstrom_fresh_2024} of explanations to the model's learned parameters and their ``\emph{faithfulness}'' \citep{wickstrom_flexibility_2025} to the model's input-output behavior. While some literature refers to both of these properties as ``faithfulness'', here we follow the naming proposed by~\citep{hedstrom_fresh_2024,wickstrom_flexibility_2025}. Both tests have been adapted to the graph-based regression task of interest in this paper, where inputs have a much larger extent than typical computer vision tasks, and the model has direct input-output skip-connections for the target's own field at t$-$6~h. Following \citep{hedstrom_fresh_2024}, we also compared all results against reference control conditions rather than relying on the inspection of raw results.

All sanity checks have been performed for the Hurricane Laura use case (see \autoref{sec:usecase}) and the following target predictions: 
\begin{itemize}
    \item[\textbf{S0}] A pool of 50 targets chosen at random from the ERA5 test set (1 Jan 2020 to 10 Jan 2021).
    \item[\textbf{S1}] The 50 targets for which the model has the highest error over the ERA5 test set.
\end{itemize} 
Due to normalization, the highest errors often occur for Vertical Velocity at 50 hPa. To have a representative pool of high-error targets across different variables, a capping mechanism was put in place: if a maximum of 10 high-error targets is reached for a given variable family, all the other high-error targets for the same variable family are ignored. The Hurricane Laura use case provides a direct sanity check of what is shown in the main text; the other two provide more robust results. This approach is replicated for IG completeness experiments in \appref{app:ig}. For all of these experiments we used the GraphCast-like trained model from \autoref{sec:usecase}.

Most of the discussion in this appendix regarding randomization and deletion sanity checks focuses on IG explanations (Fig.~\ref{fig:app:randomiz_laura}, Fig.~\ref{fig:app:deletion_laura}). However, sanity checks for IxG explanations have been performed as well (Fig.~\ref{fig:app:randomiz_robust}, Fig.~\ref{fig:app:deletion_robust}).

\subsection{Randomization tests}
\label{app:sanity_rand}

We randomized all weights at once, rather than layer-by-layer as in \citep{adebayo_sanity_2018}. The reason is to provide a general outlook of the \emph{sensitivity} of the XAI techniques implemented in the tool, rather than assessing its layer-wise granularity. Model weights are randomized with 5 different seeds. 
For each target of interest, IG input attribution maps and rankings are computed for the GraphCast-like trained model (same as in \autoref{sec:usecase}) and the 5 randomized models.

We then define 4 paired tests to compare explanations: \textbf{T-R)} explanations of trained model against those of randomized model. \textbf{R-R)} explanations of random models compared across different seeds. \textbf{T-X)} Explanation of trained model against the input. \textbf{R-X)} Explanation of randomized model against the input. The first and third (T-R and T-X) answer the questions we care about, i.e., explanation \emph{sensitivity} to model parameters (T-R) and their difference from inputs (T-X), while the other two (R-R and R-X) serve as a control. For each of these pairs two quantities are computed: Pearson's correlation ($\rho$) between the attribution tensors, and Kendall's correlation ($\tau$) between the attribution rankings. Pearson's correlation was chosen in place of Spearman correlation \citep{adebayo_sanity_2018} because the large number of almost-zero entries in the attribution tensor, which is caused by the localized receptive fields of the graph-based model, would raise Spearman correlation in a meaningless way.

\begin{figure}[p]
  \centering
  \includegraphics[width=\linewidth]{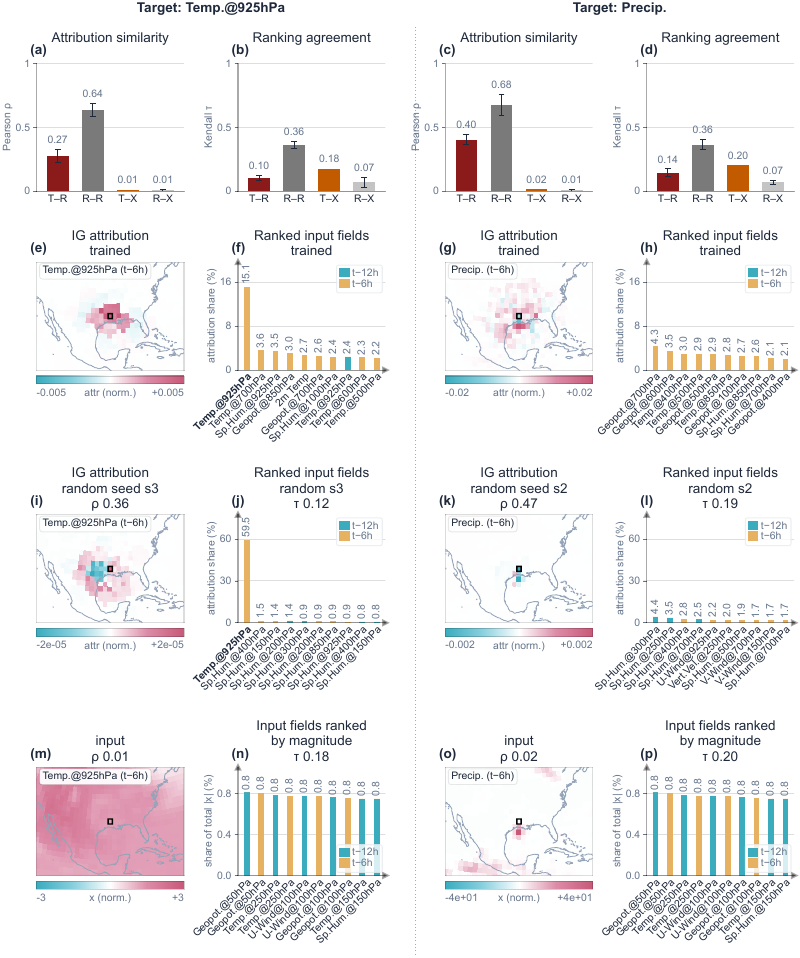}
  \caption{Randomization tests for Hurricane Laura use case in \autoref{sec:usecase}. Target XAI field is (\emph{Left}) Temperature at 925 hPa, (\emph{Right}) Precipitation. (\textbf{a,c}) Similarity of IG input attribution maps for the 4 paired tests described in \appref{app:sanity_rand} measured as Pearson's correlation. (\textbf{b,d}) Similarity of IG rankings measured as Kendall's correlation. (\textbf{Second row}) IG explanations for the Laura use case. (\textbf{Third row}) IG explanations for a model with randomized weights. For each target the seed whose IG attribution has the highest correlation with the trained model's IG attribution is shown. (\textbf{Fourth row}) Model inputs for the target's own field at t$-$6~h. Correlations with the corresponding entries in the Second row are reported.
  }
  \label{fig:app:randomiz_laura}
\end{figure}

Fig.~\ref{fig:app:randomiz_laura} shows randomization tests for the use case in Fig.~\ref{fig:explain}. While previous literature \citep{adebayo_sanity_2018} marks IG as poorly performing under randomization tests, it does perform fairly well in this specific context. Correlation of trained model's explanation with the randomized model's explanation (T-R in Fig.~\ref{fig:app:randomiz_laura}a-d) is always lower than the similarity of explanations within different randomized models (R-R in Fig.~\ref{fig:app:randomiz_laura}a-d). This holds for both input attribution maps (Fig.~\ref{fig:app:randomiz_laura}a,c) and input rankings (Fig.~\ref{fig:app:randomiz_laura}b,d). The similarity between explanations and inputs is almost null if we compare the spatial patterns (Fig.~\ref{fig:app:randomiz_laura}a,c), yet present if we compare the per-variable magnitudes (Fig.~\ref{fig:app:randomiz_laura}b,d). Input field magnitude (Fig.~\ref{fig:app:randomiz_laura}n,p) share is computed by summing per-variable absolute inputs over space and normalizing by the total over variables.

\begin{figure}[t]
  \centering
  \includegraphics[width=\linewidth]{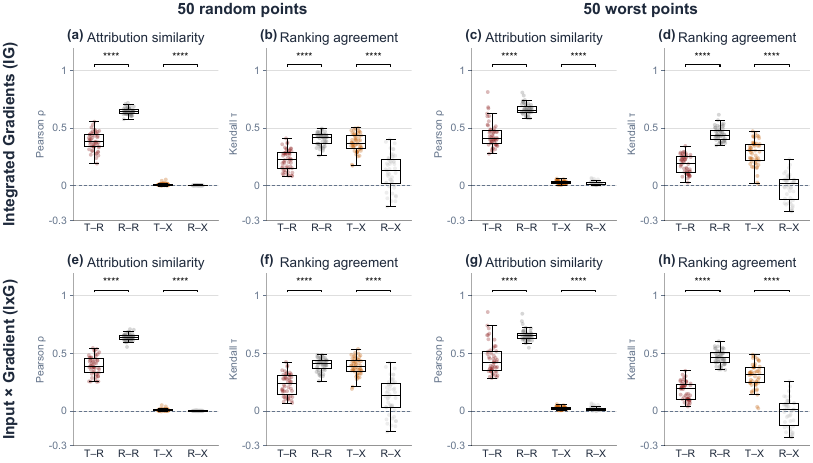}
  \caption{Randomization test for (\emph{Left}) 50 random targets (S0), and (\emph{Right}) 50 targets with highest prediction error (S1). (\textbf{First row}) IG. (\textbf{Second row}) IxG. (\textbf{Brackets}) Two-sided Wilcoxon signed-rank test, paired by target (ns $> 0.05$, * $\le 0.05$, ** $\le 0.01$, *** $\le 10^{-3}$, **** $\le 10^{-4}$). $p$-values are Bonferroni-corrected.
  }
  \label{fig:app:randomiz_robust}
\end{figure}

Fig.~\ref{fig:app:randomiz_robust}a-d shows that IG randomization test results are stable across explanations for different target predictions, and agree with what is described above for the Hurricane Laura use case.
IxG also performs well under randomization tests (Fig.~\ref{fig:app:randomiz_robust}e-h), with similar sensitivity to IG, making it a good explanation technique for fast input attributions. Nonetheless, it lacks completeness by design.

\subsection{Deletion tests}
\label{app:sanity_del}

Deletion tests ask whether the inputs that an explanation ranks highest actually drive the prediction: if they do, removing them first should change the model's output more, and sooner, than removing inputs in any other order \citep{bach_pixel-wise_2015,samek_evaluating_2017,petsiuk_rise_2018}. 

We adapted the test to our regression setting. An input is removed by setting it to zero in normalized space, i.e. to its training mean. Forcing and static features are never modified. Unlike classification tasks, the model has no class score. Hence, to quantify how much the prediction moves we measure the absolute change of the trained model's prediction at the target, $|f(x) - f(x_{\text{removed}})|$, in units of the target variable's standard deviation. Also, the target's own field at t$-$6~h at the target node is never removed: the skip-connection copies it directly to the output, so removing it would make any ranking that places it first look faithful.

For each target prediction we ran three tests, each against a control condition:
\begin{itemize}
\item [\textbf{Single-entry deletion} \citep{samek_evaluating_2017}] The $k$ input entries with the largest $|\mathrm{IG}|$ are progressively removed. This is compared with two controls: removal in random order (3 orders, averaged), and removal in order of the fields' input magnitude share (each field's sum of $|x|$ over all grid nodes, normalized by the total over the 166 fields).
\item [\textbf{Field deletion}] The $k$ input fields (one variable at one input step over all grid nodes) with the largest attribution share are progressively removed. This is compared with their removal in random order (3 orders, averaged). This tests the field ranking (Fig.~\ref{fig:explain}e-f) that is actually read in the tool, rather than single entries.
\item [\textbf{Faithfulness correlation} \citep{bhatt_evaluating_2020,bommer_finding_2024}] Each of the 166 fields alone is removed, and the resulting prediction change is recorded. Pearson's $\rho$ is then computed between these and the fields' IG shares. As a control, the same prediction changes are correlated with each field's input magnitude share, which asks whether the size of the input alone predicts which field matters.
\end{itemize}

For the single-entry and field deletion tests, the prediction change plotted against $k$ forms a most-relevant-first (MoRF) perturbation curve \citep{samek_evaluating_2017}. We summarize each per-target curve by its mean over the removal steps, i.e. the Area Over the Perturbation Curve (AOPC) \citep{samek_evaluating_2017}. The approach is adapted to our spatio-temporal regression task: the output is a regression target, so we plot the absolute change rather than the output. In this setting, AOPC measures the area under our curve. Also, since the input has $\approx 4.8$ M entries, $k$ is log-spaced, so the AOPC is the normalized area on a log-$k$ axis. $k=0$ and full removal are excluded from AOPC computation, since all orderings coincide there. 

\begin{figure}[h]
  \centering
  \includegraphics[width=\linewidth]{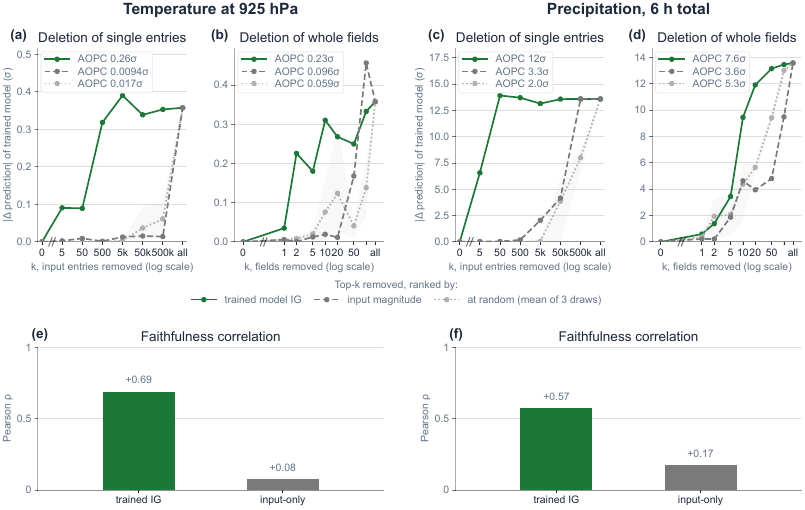}
  \caption{Deletion tests for the Hurricane Laura use case in \autoref{sec:usecase}. Target XAI field is (\emph{Left}) Temperature at 925 hPa, (\emph{Right}) Precipitation. (\textbf{a,c}) Perturbation curves when deleting single entries, ranked by (green) trained model's IG attributions, (gray) input magnitude, and (light gray) ranked at random. (\textbf{b,d}) Perturbation curves when deleting whole input fields, ranked by (green) trained model's IG attribution share, (gray) input field magnitude share, and (light gray) at random. (\textbf{e,f}) Faithfulness correlations.
  }
  \label{fig:app:deletion_laura}
\end{figure}

Fig.~\ref{fig:app:deletion_laura} shows the deletion tests described above for the use case in Fig.~\ref{fig:explain}. The MoRF perturbation curves for single entries (Fig.~\ref{fig:app:deletion_laura}a,c) and whole fields (Fig.~\ref{fig:app:deletion_laura}b,d) suggest that IG explanations well capture the inputs that drive model prediction. There seems to be some overlap between the input fields with the highest magnitude and the IG attributions (Fig.~\ref{fig:app:deletion_laura}b,d and Fig.~\ref{fig:app:randomiz_laura}b,d). Faithfulness correlations in Fig.~\ref{fig:app:deletion_laura}e-f settle this: IG reliably identifies the input fields that change model prediction the most, rather than those with the highest absolute magnitude.

\begin{figure}[t]
  \centering
  \includegraphics[width=\linewidth]{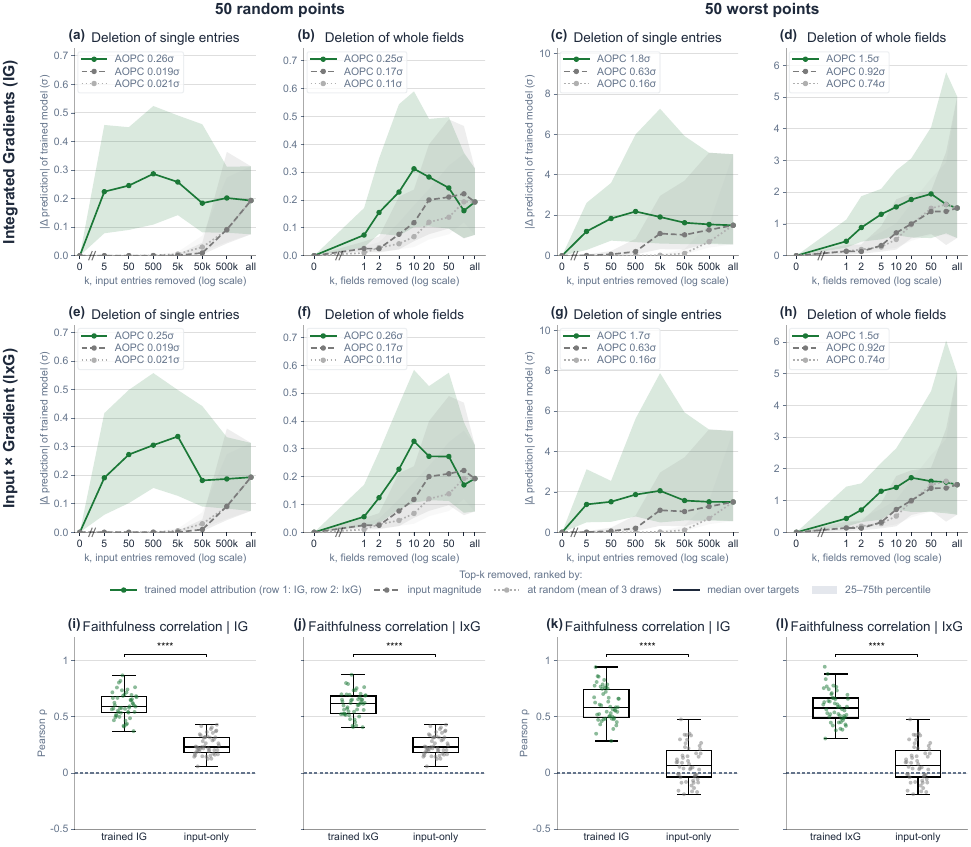}
  \caption{Deletion tests for (\emph{Left}) 50 random targets (S0), and (\emph{Right}) 50 targets with highest prediction error (S1). (\emph{First row}) IG attributions (\emph{Second row}) IxG. (\textbf{a,c,e,g}) Perturbation curves when deleting single entries, ranked by (green) trained model's IG/IxG attributions, (gray) input magnitude, and (light gray) ranked at random. (\textbf{b,d,f,h}) Perturbation curves when deleting whole input fields, ranked by (green) trained model's IG/IxG attribution share, (gray) input field magnitude share, and (light gray) at random. Lines are medians over targets, shaded areas show the range from the 25th to the 75th percentiles. (\emph{Third row}) Faithfulness correlations for (\textbf{i,k}) IG and (\textbf{j,l}) IxG. (\textbf{Brackets}) Two-sided Wilcoxon signed-rank test of trained vs.\ input-only $\rho$, paired by target (ns $> 0.05$, * $\le 0.05$, ** $\le 0.01$, *** $\le 10^{-3}$, **** $\le 10^{-4}$). $p$-values are Bonferroni-corrected.
  }
  \label{fig:app:deletion_robust}
\end{figure}

Similarly to what was done for the sanity checks, the deletion test has been conducted for the targets in S0 and S1, defined in \autoref{app:sanity_rand}. The results in Fig.~\ref{fig:app:deletion_robust} provide even stronger support to the claims in the paragraph above: IG correctly identifies the fields that drive predictions rather than relying on high-magnitude fields. IxG (Fig.~\ref{fig:app:deletion_robust}e-h,j,l) also performs well under deletion tests, supporting its use in the tool as a fast attribution method.

\section{Integrated Gradients}
\label{app:ig}

IG attribution scores are computed by integrating the gradient along the path from a reference state to the actual input. The reference state is chosen to be the climatological mean of each field. With this choice the results have to be interpreted as follows: positive values indicate features that increase the model's prediction above a baseline output value, while negative values indicate features that decrease it below the baseline output. The baseline output is the model output when its input is the climatological mean of each field, i.e. the reference input state. IG guarantees that the sum of attribution scores accounts in full for the prediction shift caused by the input moving from the baseline to the actual value of the field (this property is called completeness). 

\begin{figure}[t]
  \centering
  \includegraphics[width=\linewidth]{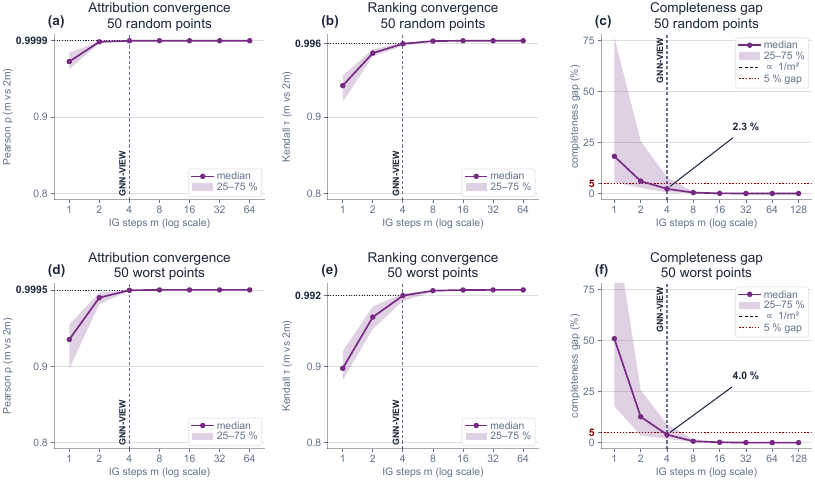}
  \caption{IG completeness tests for (\emph{First row}) 50 random targets (S0), and (\emph{Second row}) 50 targets with highest prediction error (S1). (\textbf{a,d}) Convergence of IG input attribution maps. (\textbf{b,e}) Convergence of IG input rankings. (\textbf{c,f}) Completeness gap with increasing number of integration steps ($m$).
  }
  \label{fig:app:completeness_robust}
\end{figure}

We approximate this integral with a Riemann sum, recovering a fraction (Fig.~\ref{fig:app:completeness_robust}c,f) of this shift depending on the integration steps $m$. A high number of integration steps ensures higher completeness at the cost of higher latency. 
To quantify the effects of the integration steps on the resulting attributions we computed two quantities: 1) the completeness gap (\%), expressed as the difference between the prediction shift explained by IG and the real prediction shift, divided by the latter. 2) Pearson's correlation between IG attributions computed with $m$ integration steps and twice as much ($2m$), with increasing number of integration steps $m$. 

Fig.~\ref{fig:app:completeness_robust} shows the results for the S0 and S1 target pools from \appref{app:sanity}. IG with $m=4$ steps recovers 96\% of the prediction shift at the median for the worst-points pool (S1, Fig.~\ref{fig:app:completeness_robust}f). Input attributions seem to stabilize after $m=4$: IG input attributions computed with $m=8$ integration steps are well correlated with the attributions computed with $m=4$, and the same holds for all larger $m$ (Fig.~\ref{fig:app:completeness_robust}). This holds  for both attribution maps (Fig.~\ref{fig:app:completeness_robust}a,d) and rankings (Fig.~\ref{fig:app:completeness_robust}b,e). For this reason, we choose $m=4$ as the default in the tool, to trade off latency against exactness. Nonetheless, completeness for $m=4$ still varies significantly (Fig.~\ref{fig:app:completeness_robust}c,f), at times exceeding the 5\% limit proposed by \citep{sundararajan_axiomatic_2017}. For this reason, we suggest using $m=8$ integration steps if deploying locally with GPU acceleration. 

\begin{figure}[t]
  \centering
  \includegraphics[width=\linewidth]{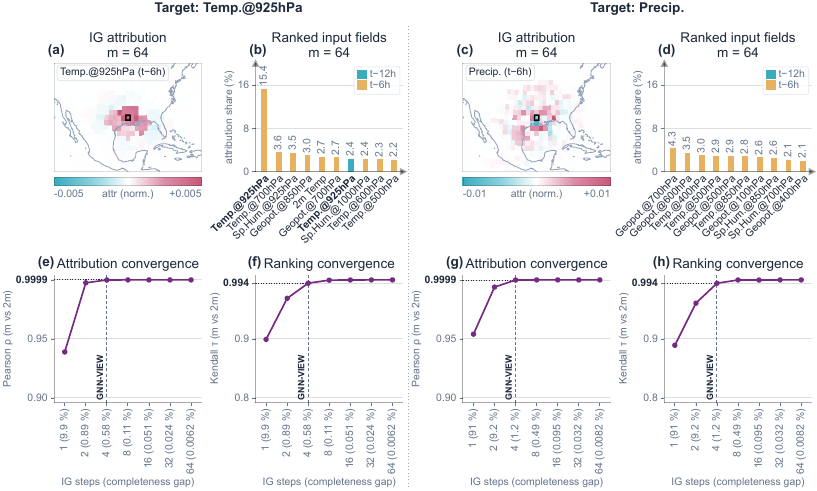}
  \caption{IG completeness tests for the Hurricane Laura use case in \autoref{sec:usecase}. Target XAI field is (\emph{Left}) Temperature at 925 hPa, (\emph{Right}) Precipitation. (\textbf{First row}) IG explanations for the Laura use case computed with $m=64$ integration steps. (\textbf{e,g}) Convergence of IG input attribution maps. (\textbf{f,h}) Convergence of IG input rankings. In all four plots the completeness gap for using a given $m$ is reported on the horizontal axis. The vertical axis reports correlations between attributions computed with $m$ and $2m$ integration steps.
  }
  \label{fig:app:completeness_laura}
\end{figure}

As for the Hurricane Laura use case discussed in \autoref{sec:usecase}, Fig.~\ref{fig:app:completeness_laura}a-d reports the same attribution panels as in Fig.~\ref{fig:explain}b,d,e,f computed with $m=64$ integration steps instead of $m=4$, as well as the convergence curves (Fig.~\ref{fig:app:completeness_laura}e-h). Using $m=64$ integration steps, IG attributions account for 99.99\% of the prediction shift, hence we can consider this as the ground truth for IG explanations.
The comparison is positive: IG with $m=4$ integration steps recovers 99.42\% of the temperature prediction shift (Fig.~\ref{fig:app:completeness_laura}e,f) and 98.8\% of the precipitation one (Fig.~\ref{fig:app:completeness_laura}g,h). The input attributions in Fig.~\ref{fig:explain} also closely resemble those in Fig.~\ref{fig:app:completeness_laura}, except for a minor reduction in magnitude of a few IG attributions in the precipitation panel (Fig.~\ref{fig:explain}d vs Fig.~\ref{fig:app:completeness_laura}c). The rankings match too, with only a change in relative position between Specific Humidity at 1000 hPa (t$-$6~h) and Temperature at 925 hPa (t$-$12~h) when going from $m=4$ to $m=64$ (Fig.~\ref{fig:explain}e vs Fig.~\ref{fig:app:completeness_laura}b). This is, however, minor if we compare their attribution share.

\end{document}